\documentclass[10pt,twocolumn,letterpaper]{article}

\usepackage[applications]{wacv}      

\usepackage{graphicx}
\usepackage{amsmath}
\usepackage{amssymb}
\usepackage{booktabs}
\usepackage{cuted}
\usepackage{multirow}

\usepackage{kotex}
\newcommand{\Tref}[1]{Table~\ref{#1}}

\newcommand{\Fref}[1]{Figure~\ref{#1}}
\usepackage[pagebackref,breaklinks,colorlinks]{hyperref}
\usepackage{pifont}

\usepackage[capitalize]{cleveref}
\crefname{section}{Sec.}{Secs.}
\Crefname{section}{Section}{Sections}
\Crefname{table}{Table}{Tables}
\crefname{table}{Tab.}{Tabs.}
\usepackage{float}
\usepackage{cite}
\usepackage{placeins}
\def\wacvPaperID{928} 
\def\confName{WACV}
\def\confYear{2025}

\title{From Street to Orbit: Training-Free Cross-View Retrieval via Location Semantics and LLM Guidance}

\author{
\centering
\begin{tabular}{c}
\text{Jeongho Min}\textsuperscript{1, *}, \text{Dongyoung Kim}\textsuperscript{2, *}, \text{Jaehyup Lee}\textsuperscript{3, $\dagger$} \\
\vspace{4pt}
\textsuperscript{1}Ulsan National Institute of Science and Technology (UNIST), South Korea \\
\textsuperscript{2}Electronics and Telecommunications Research Institute (ETRI), South Korea \\
\textsuperscript{3}Kyungpook National University, South Korea \\
\vspace{4pt}
\texttt{jeongho.min@unist.ac.kr, dongyoung.kim@etri.re.kr, jaehyuplee@knu.ac.kr}
\end{tabular}
}

\begin{document}
\maketitle

\begingroup
\renewcommand\thefootnote{}\footnotetext{%
\textsuperscript{*}Equal contribution. \quad
\textsuperscript{$\dagger$}Corresponding author.}%
\addtocounter{footnote}{-1}
\endgroup

\begin{abstract}
Cross-view image retrieval, particularly street-to-satellite matching, is a critical task for applications such as autonomous navigation, urban planning, and localization in GPS-denied environments. However, existing approaches often require supervised training on curated datasets and rely on panoramic or UAV-based images, which limits real-world deployment. In this paper, we present a simple yet effective cross-view image retrieval framework that leverages a pretrained vision encoder and a large language model (LLM), requiring no additional training. Given a monocular street-view image, our method extracts geographic cues through web-based image search and LLM-based location inference, generates a satellite query via geocoding API, and retrieves matching tiles using a pretrained vision encoder (e.g., DINOv2) with PCA-based whitening feature refinement. Despite using no ground-truth supervision or finetuning, our proposed method outperforms prior learning-based approaches on the benchmark dataset under zero-shot settings. Moreover, our pipeline enables automatic construction of semantically aligned street-to-satellite datasets, which is offering a scalable and cost-efficient alternative to manual annotation. All source codes will be made publicly available at \url{https://jeonghomin.github.io/street2orbit.github.io/}.
\end{abstract}

\section{Introduction}
\label{sec:intro}

\begin{figure}[t]
  \centering
  \includegraphics[width=\linewidth]{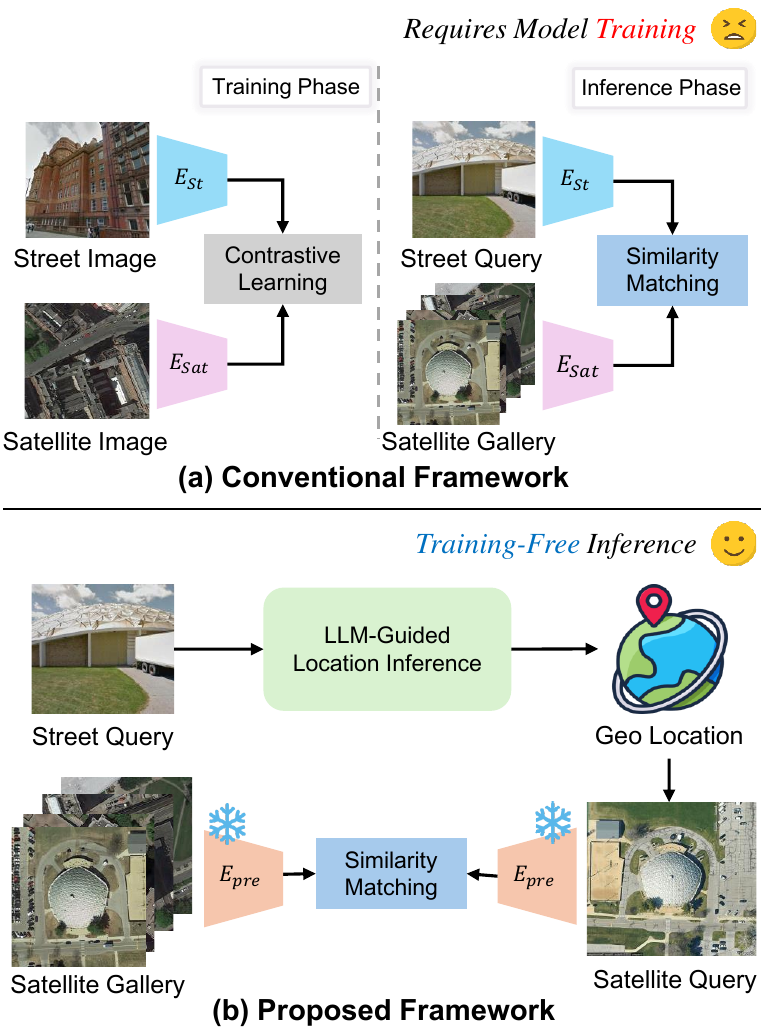}
  \vspace{-0.5cm}
    \caption{Comparison between conventional contrastive learning-based cross-view retrieval (top) and our proposed training-free, LLM-guided framework (bottom). Unlike prior work requiring separate encoders and contrastive training, our approach uses a single pretrained vision encoder (e.g., DINOv2) for both views, enabling direct retrieval in a shared feature space.}
  \vspace{-9mm}
  \label{fig:comparison}
\end{figure}

Cross-view image retrieval, especially \textit{Street-to-Satellite Retrieval}, has emerged as a critical task in various applications such as autonomous navigation, urban planning, and location-based services\cite{regmi2019bridging,toker2021coming,deuser2023sample4geo,zeng2022geo,wang2024mfrgn}. Given a street-level image, the goal is to accurately retrieve its corresponding satellite view image of the same location, despite large variations in viewpoint, scale, and appearance. This functionality is particularly valuable in GPS-denied or weak-signal environments such as urban canyons, remote areas, or disaster zones, where satellite imagery can serve as the only reliable reference for geo-localization~\cite{zeng2022geo}. 

This cross-view retrieval remains inherently challenging due to the extreme domain gap and disparity between ground and satellite viewpoints. Viewpoint distortion, lighting changes, and background clutter introduce a substantial misalignment in the visual feature space~\cite{regmi2019bridging,toker2021coming}.




To overcome these challenges, prior works adopt supervised learning approaches trained on curated datasets with strong geometric alignment (see Figure~\ref{fig:comparison} (a)).
These methods typically utilize panoramic or UAV-captured images~\cite{workman2015wide, liu2019lending,zheng2020university}, enabling dual encoders or transformer-based alignment modules to learn cross-view correspondence with the assumption that the model would be given the ideal input conditions, such as full 360-degree panoramas or low-altitude drone imagery.

Such requirements are difficult to satisfy in practice, by highly relying on ideal input assumptions and paired supervision, limiting real-world deployment. Panoramic or UAV imagery requires specialized equipment and constrained acquisition setups, making it unsuitable for general users. In contrast, monocular images from smartphones are far more accessible, lightweight, and practical for real-world deployment. They also reduce computational overhead, enabling efficient inference on mobile and edge devices. 

However, monocular photos captured in the wild present additional challenges. They often contain occlusions, poor lighting, or ambiguous structures, which degrade the performance of models trained on curated and aligned inputs. In addition, constructing large-scale, diverse paired datasets for supervised training remains costly and time-consuming.


\begin{figure*}[t]
\centering
\includegraphics[width=1.00\linewidth]{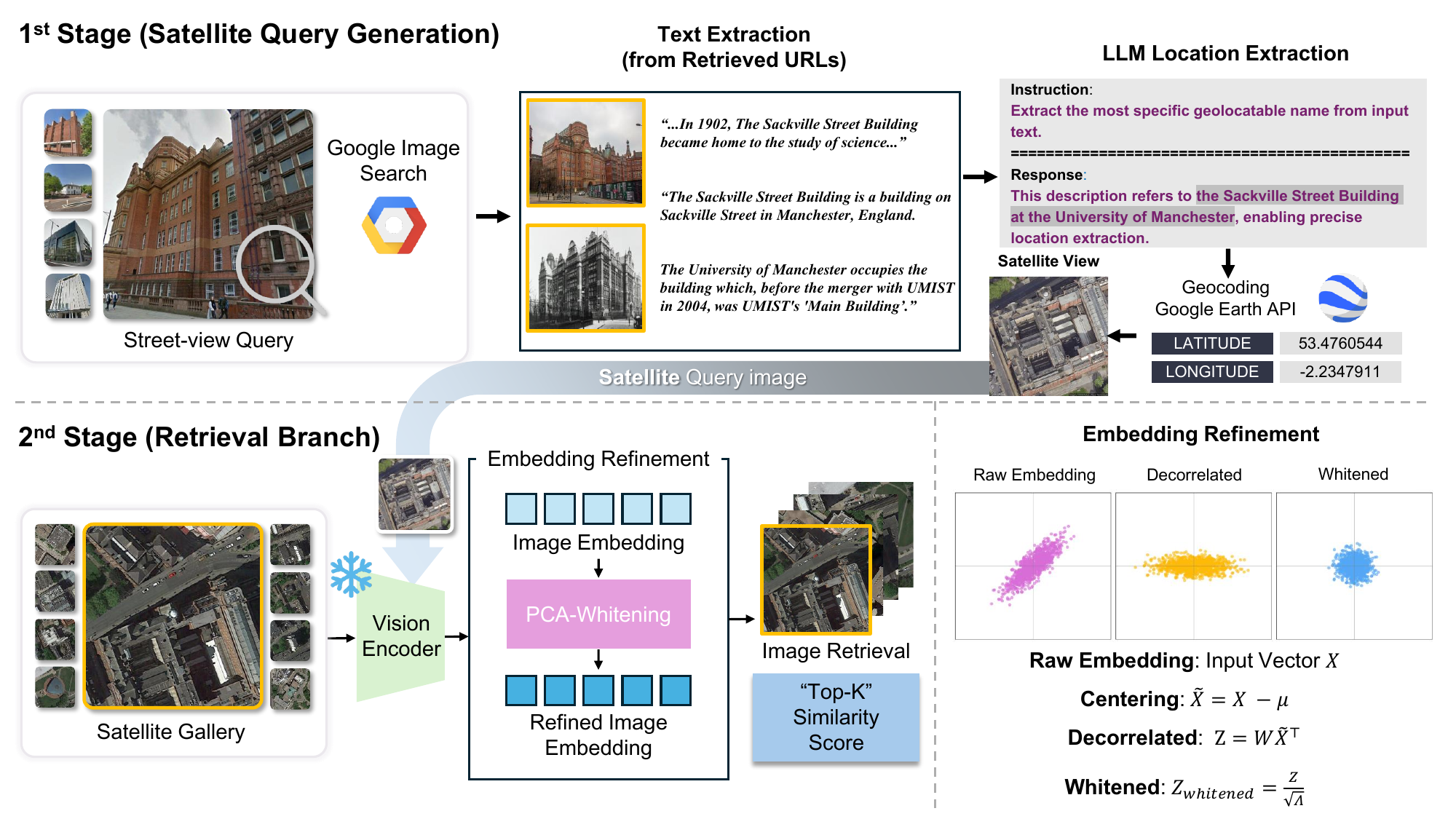}
\vspace{- 0.8 cm}
\caption{Overview of our proposed training-free cross-view retrieval pipeline. Given a street-level image, we first extract textual context using Google Image Search, then infer the most specific geolocatable name via a large language model(LLM). This name is geocoded into coordinates to retrieve a satellite tile, which is embedded using a pretrained vision encoder (e.g., DINOv2). The resulting feature is whitened and matched against a satellite gallery using similarity-based retrieval. Our method enables semantically aligned cross-view matching without any supervised training.}
\label{fig:oveview}
\vspace{-0.7cm}
\end{figure*}

To overcome these limitations, we propose a training-free, LLM-guided cross-view image retrieval framework that requires no feature-level alignment or supervised training (see Figure~\ref{fig:comparison}~(b)).
Our approach takes a single street-view photo as input, infers its geolocatable semantics using a large language model (specifically, Mistral 7B~\cite{jiang2023diego}), converts the inferred location to latitude and longitude coordinates through geocoding, and generates the corresponding satellite query via map APIs from a pre-indexed satellite gallery.
The retrieved satellite image is then encoded using a large-scale pretrained vision encoder~\cite{dosovitskiy2020image, radford2021learning, caron2021emerging, zhai2023sigmoid} without any additional fine-tuning, 
while PCA-whitening is applied to suppress low-level artifacts such as lighting and texture biases in the embedding space, thereby improving retrieval robustness~\cite{radenovic2018fine, babenko2015aggregating, zhang2024image}.
Through extensive experiments, we demonstrate that this embedding refinement consistently improves zero-shot performance not only in the satellite domain but also for general image retrieval tasks.

Moreover, our framework is designed to flexibly leverage spatial metadata, coordinate systems, and geocoding APIs, making it highly practical and generalizable for unconstrained monocular street photos in real-world scenarios.
In particular, our method achieves state-of-the-art performance in the challenging Street-to-Satellite retrieval task of the University-1652 benchmark~\cite{zheng2020university}, despite relying solely on single-view queries without drone imagery or paired supervision.
Furthermore, the proposed pipeline naturally enables automatic construction of Street-to-Satellite image pairs, allowing scalable dataset generation without manual annotation. This provides a practical foundation to overcome the limitations of existing benchmarks and facilitates broader data accessibility for future learning-based cross-view research.



\vspace{2mm}
\noindent\textbf{Our main contributions are summarized as follows:}
\begin{itemize}\setlength{\itemsep}{2pt}

  \item We present a training-free and real-world photo compatible retrieval framework that infers location semantics from monocular street-view images using LLMs, without requiring paired supervision. 
  \item We introduce a lightweight embedding refinement using PCA-whitening, which improves retrieval robustness in zero-shot settings. 
  \item Our method achieves state-of-the-art performance on University-1652, surpassing existing models under both zero-shot and retrieval-based localization settings, while also enabling scalable dataset construction through automatic street-satellite pairing.
\end{itemize}


\begin{figure*}[t]
\centering
\includegraphics[width=0.9\linewidth]{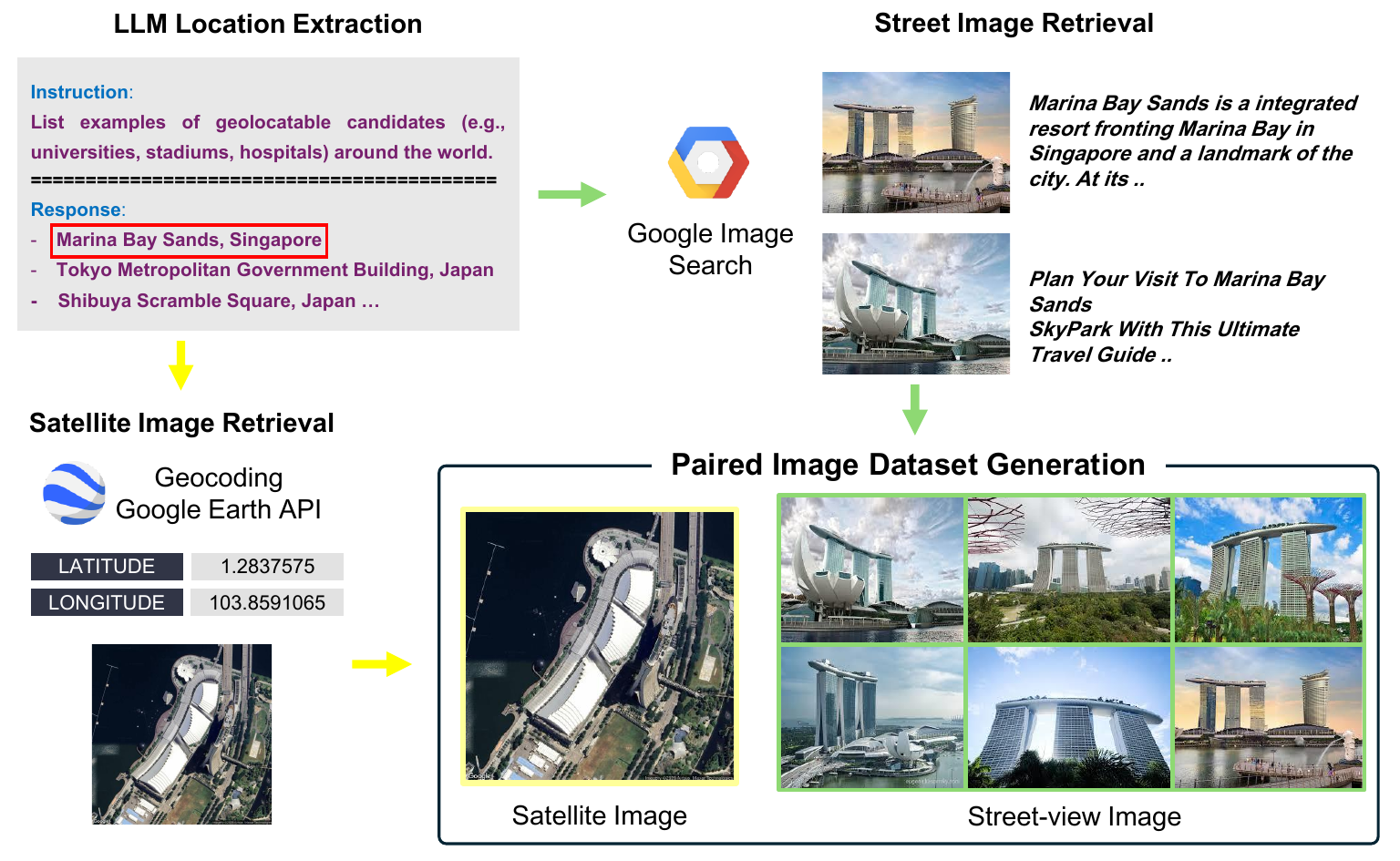}
\vspace{- 0.3 cm}
\caption{Overview of our dataset generation pipeline. We prompt an LLM to produce a list of globally recognizable locations (e.g., ``Marina Bay Sands", ``Tokyo Metropolitan Government Building"). For each location, we retrieve street-view images via Google Image Search and obtain geographic coordinates using a Google Geocoding API. These coordinates are used to generate corresponding satellite tiles through the Google Static Maps API. This process produces high-quality, semantically aligned street-to-satellite image pairs at scale. }
\label{fig:generation}
\vspace{- 0.4 cm}

\end{figure*}


\vspace{- 0.2 cm}
\section{Related Works}
\subsection{Cross view image retrieval}
Cross-view image retrieval aims to retrieve images of the same location captured from drastically different perspectives, typically street-level views against top-down satellite images.
To tackle this challenging task, prior studies have proposed various methods that utilize cross-view visual features to retrieve the most semantically relevant image pairs~\cite{wang2021each, wang2024mfrgn, deuser2023sample4geo, ye2024cross}.

Early efforts modeled the problem using Siamese CNN architectures~\cite{koch2015siamese} that minimized feature-space distances between paired ground-to-satellite images~\cite{liu2019lending, tian2020cross, shi2019spatial}. However, the drastic domain shift, which includes viewpoint distortions, occlusions, and inconsistent scale, limits the performance of purely CNN-based models. 

To address this, LPN~\cite{wang2021each} proposed part-based attention mechanisms that divide ground-level inputs into spatial regions and aggregate them using learned weights, effectively emphasizing informative subregions. While this improved feature discriminability, the fundamental gap between ground and aerial perspectives remained substantial. 

Subsequent work introduced intermediate UAV imagery to bridge the domain gap. For instance, PLCD~\cite{zeng2022geo} jointly trains on drone-to-satellite and ground-to-satellite retrieval tasks, treating drone views as a natural intermediary. This tri-view strategy reduces the viewpoint discrepancy and improves retrieval robustness. 

Beyond tri-view architectures, drone-centric retrieval frameworks have recently emerged as a distinct research direction that fully substitutes street-level inputs with UAV imagery. 
DWDR~\cite{wang2024learning} enhances robust matching by using diverse rotation augmentations and local patches of Drone images. MCCG~\cite{shen2023mccg} enforces cross-view consistency through a multi-classifier scheme based on ConvNeXt backbones, while Sample4Geo~\cite{deuser2023sample4geo} simplifies the retrieval pipeline by adopting symmetric contrastive learning using GPS-aligned positives and visually hard negatives, without relying on polar transformations or dense alignment.

However, retrieving corresponding satellite views from a single, narrow field-of-view street image remains extremely challenging due to severe viewpoint and scale disparities. To mitigate this, recent studies such as MFRGN~\cite{wang2024mfrgn} and Panorama-BEV~\cite{ye2024cross} attempt to bridge the domain gap by leveraging panoramic street images or transforming them into Bird’s-Eye View (BEV) representations. These methods reduce the perspective discrepancy by expanding the field of view or generating intermediate geometric representations better aligned with satellite perspectives.

Although effective, these methods typically assume access to panoramic inputs, low-altitude drone imagery, or well-aligned assumptions about well-aligned training data that often do not hold in practical scenarios. 
In contrast, our approach embraces single-view, in-the-wild monocular photos and removes the need for supervised training or explicit geometric conversion, making it more scalable and applicable to real-world deployments.

\subsection{Image Encoder}
Progress in vision-language modeling and self-supervised learning has led to powerful image encoders that generalize well across domains and tasks. In particular, transformer-based encoders\cite{dosovitskiy2020image, liu2021swin, liu2022swin, caron2021emerging} have shown strong capabilities in learning semantically rich representations. 

ViT (Vision Transformer)~\cite{dosovitskiy2020image} processes images as tokenized patches and models long-range dependencies via self-attention. These global features make ViT-based backbones a natural fit for high-level retrieval and recognition tasks. CLIP~\cite{radford2021learning} introduces contrastive learning over image-text pairs at scale, resulting in a joint embedding space that supports zero-shot classification and retrieval. However, its effectiveness deteriorates in domains like satellite or aerial imagery, where the distribution of content diverges significantly from that of natural images. 

To bridge this gap, RemoteCLIP~\cite{liu2024remoteclip} retrains the CLIP architecture on domain-specific satellite and UAV datasets, improving both semantic fidelity and alignment in remote sensing contexts. 
In addition, DINO~\cite{caron2021emerging} introduces a ViT-based self-distillation framework that iteratively transfers knowledge between student and teacher networks, yielding robust frozen features that generalize well across tasks.
Building on this, DINOv2~\cite{oquab2023dinov2} scales up the training with a larger curated dataset and incorporates techniques such as SwAV~\cite{caron2020unsupervised} centering and the KoLeo regularizer~\cite{sablayrolles2018spreading} to achieve semantic representations comparable to weakly-supervised models, despite being purely self-supervised. 
Meanwhile, SigLIP~\cite{zhai2023sigmoid} modifies the CLIP architecture by replacing the contrastive loss with a binary classification loss, mitigating label imbalance and improving generalization to diverse image-text pairs.
SigLIP v2~\cite{tschannen2025siglip} extends this framework with deeper ViT backbones, longer pretraining, and fine-grained tuning, achieving stronger multimodal zero-shot performance.

In this work, we leverage these state-of-the-art pretrained vision encoders to encode satellite imagery in a training-free setup. Its frozen representations, combined with coordinate-guided search and language-driven semantic inference, allow for robust retrieval in the absence of paired supervision or domain-specific fine-tuning.

\section{Method}
We propose a novel training-free, language-guided cross-view retrieval framework designed to handle real-world monocular photos without requiring any paired supervision. As illustrated in Figure~\ref{fig:oveview}, our pipeline consists of three main components: (i) Satellite query Generation via location semantics, (ii) Visual Embedding and Similarity-based Retrieval, and (iii) Embedding Refinement for robust matching. Furthermore, this pipeline naturally enables large-scale Photo-to-Satellite dataset construction with minimal manual effort. 


\subsection{Satellite Query Image Generation via Location Semantics}
\label{method3-1}
\noindent\textbf{Location-Centric Semantic Extraction}
Given an unconstrained street-view image, we leverage the Google Image Search interface to retrieve surrounding contextual information such as page metadata, descriptive captions, text content, and related keywords. This multimodal signal forms a noisy but information-rich description from which geospatial semantics can be inferred. For batch queries, descriptive texts across all candidate scenes are concatenated into a single prompt to increase contextual coherence. At this time, url's metadata, web document body text, and text information around the image collected from the search result are used together. If there are several Image queries, the information corresponding to each photo scene is collected in the description at once.

\noindent\textbf{LLM-based Location Inference }
To extract the most specific geolocatable name (e.g., landmark, buildings, or administrative regions), we prompt a large language model~\cite{guo2025deepseek,achiam2023gpt,touvron2023llama} with the aggregated textural context. The prompt is designed to prioritize physical place names over abstract or event-based descriptors. This LLM-based inference effectively filters out irrelevant noise while capturing semantically meaningful references suitable for geocoding. 

We employ \textbf{Mistral 7B}~\cite{jiang2023diego} for this task due to its strong performance on reasoning and language understanding benchmarks, surpassing even larger models such as LLaMA 2 13B and 34B~\cite{touvron2023llama} across tasks like commonsense QA, math, and reading comprehension. Additionally, its use of grouped-query attention (GQA) and sliding window attention (SWA) ensures fast inference and efficient memory usage—an essential factor in real-time geolocation pipelines. As an open and lightweight model, Mistral 7B offers an optimal balance of performance, efficiency, and deployability in practical applications.

This step plays a critical role in disambiguating verbose or indirect location mentions. For example, as illustrated in Figure~\ref{fig:oveview}, a complex input such as:

\begin{quote}
\vspace{-0.1cm}
``In 1902, The Sackville Street Building became home to the study of science..." \\
``The Sackville Street Building is a building on Sackville Street in Manchester, England. ..."
\vspace{-0.1cm}
\end{quote}

is interpreted by the LLM as:

\begin{quote}
\vspace{-0.1cm}
\textit{“This description refers to the \textbf{Sackville Street Building} at the University of Manchester, enabling precise location extraction.”}
\vspace{-0.1cm}
\end{quote}

Such inference allows our system to confidently retrieve accurate geocoordinates via geocoding APIs, even when the input descriptions are indirect or historical. Through this semantic understanding, we ensure that the generated satellite queries faithfully reflect the true geographic intent behind the original image or description.


\noindent\textbf{Geocoding and Satellite Image Generation.}
The location string inferred by the LLM is submitted to the Google Maps Geocoding API to convert it into geographic coordinates (i.e., latitude and longitude). Based on these coordinates, we retrieve a satellite image centered at the inferred location via Google Earth Engine and the Google Static Maps API with a fixed zoom level and resolution. The resulting satellite tile serves as the retrieval query proxy that bridges the domain gap between the original street-view image and satellite-view, and is used as input in the retrieval branch. 


\subsection{Retrieval Branch with Visual Embedding and Similarity Matching}
\noindent\textbf{Feature Encoding and Retrieval.}
The satellite query image, along with a pre-indexed satellite gallery, is processed using the same frozen vision encoder. In our experiments, we adopt DINOv2 as the backbone, extracting 768-dimensional global features. For retrieval, cosine similarity is computed between the query embedding and gallery entries, and top-$k$ nearest candidates are selected based on similarity scores. 


\noindent\textbf{Embedding Refinement via Whitening.}
Although DINOv2 provides semantically rich features, raw embeddings may still be affected by low-level artifacts such as illumination or texture biases. To address this issue, we newly introduce to apply a lightweight PCA-whitening procedure at inference time. Specifically, given an embedding matrix \( X \in \mathbb{R}^{n \times d} \), we perform mean-centering: 
\vspace{-0.1cm}

\[
\tilde{X} = X - \mu,
\]
\vspace{-0.1cm}

followed by projection via the top-$k$ principal components \( W \in \mathbb{R}^{k \times d} \):
\vspace{-0.3cm}

\[
Z = W \tilde{X}^\top.
\]
\vspace{-0.1cm}

and final whitened normalization: 
\vspace{-0.1cm}

\[
Z_{\text{white}} = \Lambda^{-\frac{1}{2}} Z,
\]
where \( \Lambda \in \mathbb{R}^{k \times k} \) contains the principal eigenvalues. The resulting representation \( Z_{\text{white}} \) consists of decorrelated components with unit variance.  

\subsection{Scalable Photo-to-Satellite Dataset Construction }
As illustrated in \Fref{fig:generation}, our framework enables automatic construction of aligned street-to-satellite pairs by leveraging LLM-based scene inference and satellite tile fetching. Given a set of seed place names (e.g., ``Marina Bay Sands", ``Colosseum"), we use the Google Image API to collect diverse real-world street-view photos. These are then paired with satellite images generated via the same geocoding pipeline described in Section \ref{method3-1}.

Note that this automated pairing process not only removes the need for human annotation but also supports scalable expansion across diverse regions. The resulting dataset contains semantically aligned image pairs that can serve as supervision for future learning-based models, enabling broader generalization beyond limited benchmark regions.




\begin{table}[t]
    \centering
    \huge
    \resizebox{\columnwidth}{!}{%
    \begin{tabular}{lccccc}
        \toprule
        \textbf{Model} & \textbf{Drone} & \textbf{R@1} & \textbf{R@5} & \textbf{R@10} & \textbf{R@1\%} \\
        \midrule
        LPN~\cite{wang2021each} & \ding{55} & 0.43 & 1.59 & 2.87 & 3.02 \\
        LPN~\cite{wang2021each} & \ding{51} & 1.28 & 3.84 & 6.59 & 6.98 \\
        Swin-B + DWDR~\cite{liu2021swin,wang2024learning} & \ding{55} & 0.35 & 2.33 & 3.88 & 4.30 \\
        MCCG~\cite{shen2023mccg} & \ding{55} & 0.93 & 3.30 & 6.01 & 6.55 \\
        Sample4Geo~\cite{deuser2023sample4geo} & \ding{55} & 1.35 & 4.76 & 7.63 & 8.02 \\
        PLCD~\cite{zeng2022geo} & \ding{51} & 6.86 & 14.39 & 18.50 & 19.15 \\
        \midrule
        Ours (DINOv2-Large) & \ding{55} & 22.84 & 33.48 & 37.93 & 37.64 \\
        \textbf{+ Refinement (256D)} & \ding{55} & \textbf{25.57} & \textbf{37.21} & \textbf{40.66} & \textbf{39.94} \\
        \bottomrule
    \end{tabular}%
    }
    \caption{\textbf{Retrieval performance on University-1652 (Street$\rightarrow$Satellite).} Top-$k$ recall metrics are reported. `\ding{51}` indicates drone images were used in training, `\ding{55}` indicates only ground-satellite views were used.}
    \label{tab:uni_retrieval}
    \vspace{-5mm}
\end{table}

\section{Experiments}
\subsection{Dataset}
We evaluate out method on the University-1652 benchmark~\cite{zheng2020university}, a large-scale dataset designed for cross-view geo-localization. It contains 1,652 buildings from 72 global universities, captured from three viewpoints: street, drone, and satellite. The training dataset comprises 701 buildings from 33 universities, while the remaining 951 buildings from 39 universities are used for test, ensuring no overlap. 

Each building is associated with street-view images collected from Google Street View and web image search, drone-view images captured along spiral flight trajectories, and high-resolution satellite images. Among the supported retrieval tasks, we focus on the most challenging scenario: Street-to-Satellite retrieval, which involves matching a monocular ground-level image to its corresponding satellite view under significant viewpoint and modality shifts. 


\subsection{Implementation Details}
In this study, we use web-based automation (Selenium) to collect relevant textual metadata from Google Image Search and apply LLM-based location inference using Mistral 7B (v0.3)~\cite{jiang2023diego}.
The extracted place names are then converted into geo-coordinates via the Google Geocoding API (v1), and satellite query images are generated centered on these coordinates from Google Static Maps API (v1).

For the retrieval stage, we utilize pretrained vision encoders (DINOv2, CLIP, RemoteCLIP, etc.) and perform inference without any additional fine-tuning.
All experiments were conducted using only the first image in each query set, assuming one image per folder setting. All embeddings are further normalized through a simple PCA-whitening process and used for final similarity-based matching.

For fair comparison, baseline models based on contrastive learning (LPN, DWDR, MCCG, Sample4Geo, PLCD) were reproduced using the official PyTorch implementations and hyperparameters provided in each paper.
All experiments were conducted on a single NVIDIA RTX A6000 GPU under the same hardware environment.

\subsection{Evaluation Metric}
To evaluate retrieval performance, we use the standard \textbf{Recall@}${k}$ metric. Given a query image, Recall@${k}$ measures the fraction of queries for which a relevant image is retrieved within the top-$k$ candidates:





Recall@${k}$ is defined as follows:
\vspace{-0.1cm}
\begin{equation}
    \mathrm{R@}k = \frac{\mathrm{Number\ of\ relevant\ items\ in\ top}\ k}{\mathrm{Total\ number\ of\ relevant\ items}}
    \label{eq:recallk}
\end{equation}

In this equation, the numerator represents the number of relevant items retrieved in the top-$k$ results for each query, and the denominator denotes the total number of relevant items for that query. By varying $k$, Recall@${k}$ provides a comprehensive view of the retrieval system’s performance under different ranking thresholds. In this paper, we report \textbf{R@1}, \textbf{R@5}, \textbf{R@10}, and \textbf{R1\%} as main evaluation metrics.

\begin{table}[t]
    \centering
    \normalsize
    \resizebox{\columnwidth}{!}{%
    \begin{tabular}{lcccc}
        \toprule
        \textbf{Model} & \textbf{R@1} & \textbf{R@5} & \textbf{R@10} & \textbf{R@1\%} \\
        \midrule
        Remote-CLIP (ViT-B/32) & 2.30 & 6.03 & 9.20 & 8.19 \\
        Remote-CLIP (ViT-L/14) & 2.44 & 5.89 & 7.90 & 7.61 \\
        Remote-CLIP (RN-50) & 1.87 & 5.60 & 8.76 & 8.05 \\
        \midrule
        CLIP (ViT-B/32) & 7.18 & 15.37 & 19.68 & 19.40 \\
        CLIP (ViT-L/14) & 10.49 & 22.99 & 30.03 & 28.59 \\
        CLIP (RN-50) & 4.31 & 11.49 & 16.09 & 15.37 \\
        \midrule
        DINOv2-Base & 22.27 & 32.33 & 36.93 & 36.64 \\
        \ + Refinement (512D) & \textbf{26.01} & 34.20 & 35.34 & 35.20 \\
        \ + Refinement (256D) & 23.28 & \textbf{33.62} & \textbf{37.93} & \textbf{37.21} \\
        \midrule
        DINOv2-Large & 22.84 & 33.48 & 37.93 & 37.64 \\
        \ + Refinement (512D) & \textbf{26.29} & 35.92 & 39.80 & 39.08 \\
        \ + Refinement (256D) & 25.57 & \textbf{37.21} & \textbf{40.66} & \textbf{39.94} \\
        \bottomrule
    \end{tabular}%
    }
    \caption{\textbf{Retrieval performance on University-1652 (Street$\rightarrow$Satellite).} Top-k and Top-1\% recall are reported. Our method, without any supervised training, achieves state-of-the-art results, and the PCA-whitening refinement ``Refinement" further boosts performance across various pretrained encoders.}
    \label{tab:uni_full}
    \vspace{-5mm}
\end{table}

\begin{table*}[t]
    \centering
    \setlength{\tabcolsep}{3pt}
    \begin{tabular}{l|cccc|cccc}
        \toprule
        \multirow{2}{*}{\textbf{Model}} & \multicolumn{4}{c|}{\textbf{ILIAS}} & \multicolumn{4}{c}{\textbf{University-1652 (Street$\rightarrow$Satellite)}} \\
        \cmidrule(lr){2-5} \cmidrule(lr){6-9}
        & \textbf{R@1} & \textbf{R@5} & \textbf{R@10} & \textbf{R@1\%} & \textbf{R@1} & \textbf{R@5} & \textbf{R@10} & \textbf{R@1\%} \\
        \midrule
        ViT-base & 28.10 & 42.00 & 49.60 & 68.40 & 5.42 & 12.70 & 18.12 & 17.26 \\
        + Refinement (D-512) & \textbf{30.80} & \textbf{47.00} & \textbf{54.40} & \textbf{70.60} & \textbf{9.27} & 19.54 & 23.82 & 23.11 \\
        + Refinement (D-256) & 28.40 & 44.00 & 52.50 & 69.80 & 7.99 & \textbf{19.69} & \textbf{24.96} & \textbf{23.25} \\
        \midrule
        CLIP-base & 47.70 & 67.60 & 74.40 & 84.40 & 13.12 & 24.11 & \textbf{31.95} & \textbf{30.67} \\
        + Refinement (D-512) & 53.20 & 73.10 & 80.20 & 90.10 & 11.13 & 19.12 & 24.25 & 23.54 \\
        + Refinement (D-256) & \textbf{57.30} & \textbf{76.10} & \textbf{82.50} & \textbf{91.80} & \textbf{14.27} & \textbf{26.53} & 31.24 & \textbf{30.67} \\
        \midrule
        DINOv2-base & \textbf{54.70} & 68.90 & 75.10 & 85.60 & 12.84 & 24.82 & 30.96 & 30.10 \\
        + Refinement (D-512) & 53.00 & \textbf{70.00} & \textbf{77.30} & \textbf{87.00} & \textbf{13.27} & 24.68 & 30.39 & 29.67 \\
        + Refinement (D-256) & 51.30 & 67.60 & 74.90 & 86.80 & 13.12 & \textbf{26.68} & \textbf{34.38} & \textbf{32.95} \\
        \midrule
        SigLIP-base & 68.90 & 83.60 & \textbf{88.30} & \textbf{94.10} & 11.41 & 20.40 & 29.24 & 27.96 \\
        + Refinement (D-512) & \textbf{70.40} & 83.20 & 87.50 & 93.50 & 13.69 & 24.54 & 30.81 & 29.96 \\
        + Refinement (D-256) & \textbf{70.40} & \textbf{83.70} & \textbf{88.30} & \textbf{94.10} & \textbf{14.12} & \textbf{27.25} & \textbf{34.52} & \textbf{32.67} \\
        \midrule
        SigLIP2-base & \textbf{68.80} & 81.50 & 87.10 & 94.10 & 19.54 & \textbf{37.09} & 44.94 & 43.79 \\
        + Refinement (D-512) & 68.10 & 80.80 & 85.50 & 92.40 & 18.40 & 31.67 & 37.38 & 36.66 \\
        + Refinement (D-256) & 68.40 & \textbf{83.50} & \textbf{88.40} & \textbf{94.90} & \textbf{20.97} & 36.52 & \textbf{45.22} & \textbf{44.51} \\
        \bottomrule
    \end{tabular}
    \vspace{-0.1cm}
    \caption{\textbf{Ablation study on ILIAS and University-1652 datasets.} We evaluate the effect of embedding refinement (PCA-whitening), input variation, and encoder choices on retrieval and localization performance. Top-$k$ retrieval and Top-1\% localization accuracy are reported. \textbf{Bold} highlights the best result in each block.}
    \label{tab:ablation}
    \vspace{-4mm}
\end{table*}
\subsection{Cross-view Model Evaluation Results}


\Tref{tab:uni_retrieval} shows comparison of the retrieval results of the Street-to-Satellite on the University-1652 benchmark. Our proposed training-free framework achieves a Top-1 recall of \textbf{22.84\%} using DINOv2-Large, outperforming several supervised methods such as LPN~\cite{wang2021each}, DWDR~\cite{wang2024learning}, MCCG~\cite{shen2023mccg}, and Sample4Geo~\cite{deuser2023sample4geo}, which rely on curated training data and a complex alignment process. 

Notably, PLCD~\cite{zeng2022geo}, which leverages drone images as an intermediate view, achieves 6.86\% Top-1 accuracy. In contrast, our method surpasses it by more than 3$\times$, without relying on drone input or any additional training, underscoring the strength of our language-guided query generation and pretrained encoder.
This demonstrates that our method can outperform existing supervised approaches even without auxiliary views such as drones.

In addition, applying our PCA-whitening refinement (dimensional reduction from 768 to 256) further boosts the Top-1 score to 25.57, demonstrating that decorrelating and normalizing features enhance retrieval robustness in zero-shot conditions. Overall, these results suggest that our training-free framework provides a practical and efficient solution for cross-view retrieval, well-suited for real-world applications.




\begin{figure*}[h]
  \centering
  \includegraphics[width=\linewidth]{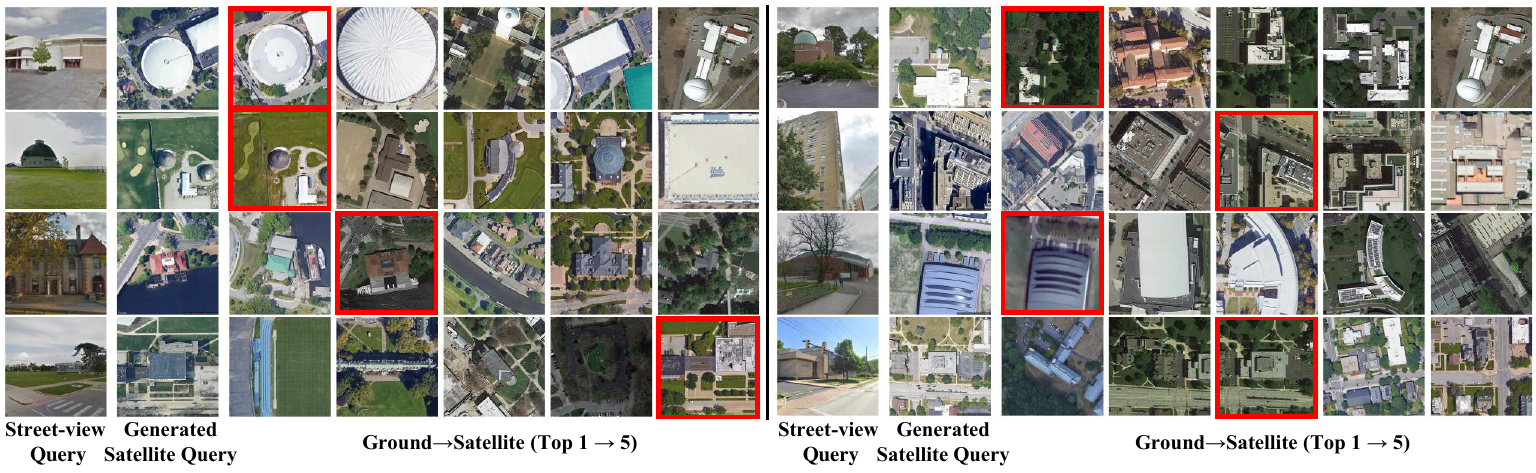}
  \vspace{-0.8cm}
  \caption{Qualitative results of our training-free retrieval framework. Given a street-level query image (left), our method retrieves the most semantically aligned satellite tile (right) using a pipeline combining Google Image Search, an LLM-based geolocation inference, geocoding, and DINOv2 with PCA-whitening. The red bounding boxes in each satellite image indicate the correctly retrieved region corresponding to the inferred location.  The top row shows accurate retrieval of a domed building with distinctive geometry; the second row demonstrates robustness in a greenery scene with minimal context, such as a house; the third row highlights generalization to an urban street with ambiguous visual representations. These samples illustrate our proposed framework's ability to perform reliable cross-view matching across diverse environments, without any kind of supervised training.}
  \label{fig:qualitative}
  \vspace{-0.5cm}
\end{figure*}
\subsection{Location Inference Accuracy}

The first stage of our proposed framework infers geo-coordinates from a street-view by extracting semantic cues with an LLM and converting them into latitude-longitude pairs via a geocoding API. We assess this step using 700 queries from University-1652, measuring distance between inferred and GT coordinates with Haversine formula~\cite{van2012heavenly}. Accuracy is reported across multiple distance thresholds. 

As summarized in Table~\ref{tab:stage1_location_accuracy}, the first stage achieves 54.75\% accuracy within 0.5 km and 65\% within 5 km. These results demonstrate that the first stage of our framework provides reliable localization performance even without any visual matching.

\begin{table}[!]
    \centering
    \huge
    \resizebox{\columnwidth}{!}{%
    \begin{tabular}{lcccc}
        \toprule
        \textbf{Distance Threshold (km)} & \textbf{0.5 km} & \textbf{1.0 km} & \textbf{2.0 km} & \textbf{5.0 km} \\
        \midrule
        \textbf{Accuracy (\%)} & 54.57 & 60.57 & 64.14 & 65.00 \\
        \textbf{Matched Samples (out of 700)} & 382 & 424 & 449 & 455 \\
        \bottomrule
    \end{tabular}%
    }
    \vspace{-0.2cm}
    \caption{\textbf{The first stage location inference accuracy.} Percentage of queries whose inferred coordinates fall within the specified distance thresholds from the GT in University-1652.}
%
    \label{tab:stage1_location_accuracy}
    \vspace{-6mm}
\end{table}



\subsection{Evaluation Across Diverse Visual Encoders}
To examine the generalizability of our approach, we evaluate multiple vision encoders under the same retrieval framework, as shown in~\Tref{tab:uni_full}. The compared encoders include CLIP~\cite{radford2021learning}, RemoteCLIP~\cite{liu2024remoteclip}, and DINOv2~\cite{oquab2023dinov2}. CLIP-based models perform moderately well, with ViT-L/14 outperforming other CLIP variants, including ResNet-50 and ViT-B/32. Interestingly, RemoteCLIP, despite its domain-specific pretraining, underperforms compared to CLIP, suggesting limited cross-view transferability. 

In contrast, self-supervised vision encoders such as DINOv2 show superior performance compared to CLIP-based models. Additionally, as shown in~\Tref{tab:uni_full}, embedding refinement using PCA-whitening consistently improves performance for both DINOv2-Base and DINOv2-Large models through dimensionality reduction. Especially, the Top-1\% score increases up to 39.94, indicating enhanced similarity matching and improved domain generalization in retrieval tasks. These results demonstrate that our framework can be effectively applied across different encoders and consistently improves performance in a training-free manner.

\subsection{Ablation Study}

Table~\ref{tab:ablation} presents the ablation study results that validate the effectiveness of the proposed PCA-whitening embedding refinement. We conduct experiments on both a general natural image retrieval dataset, \textbf{ILIAS}~\cite{kordopatis2025ilias}, and a cross-view geo-localization benchmark, \textbf{University-1652}~\cite{zheng2020university}, which focuses on the UAV$\rightarrow$Satellite retrieval scenario.

The results show that applying our simple refinement consistently improves retrieval performance across most pretrained vision encoders. Notably, the benefit is more pronounced in \textbf{University-1652} benchmark, which involves a significant domain gap between street and satellite views. For example, DINOv2-base encoder achieves a Top-1 Recall of 12.84\% without refinement, increasing to 13.27\% and 13.12\% with refinement (D-512, D-256), respectively.


A similar trend is observed on the general retrieval benchmark \textbf{ILIAS}. For instance, CLIP-base improves from 47.70\% Top-1 Recall without refinement to 57.30\% with refinement (D-256). Even strong basedline encoders, such as SigLIP and SigLIP2, maintain or slightly improve their Top-1 accuracy above 70\% with refinement applied.

These findings suggest that our simple PCA-whitening embedding refinement effectively stabilizes feature representations in zero-shot scenarios, yielding consistent performance gains not only for complex cross-view geo-localization tasks but also for standard natural image retrieval pipelines, without any additional training.


\subsection{Qualitative Results}



\Fref{fig:qualitative} presents qualitative Top-5 retrieval results of our proposed training-free framework applied to the University-1652 dataset.
For each street-view query image, our method first infers the location name using an LLM and retrieves the corresponding satellite query through geocoding.
The generated query is then matched against the satellite gallery in the retrieval branch, and the Top-$k$ most similar candidates are retrieved based on the refined DINOv2 embeddings.

Due to the nature of our LLM-based satellite query generation pipeline, there may be cases where the generated query slightly differs from the ground-truth satellite image in the dataset.
For example, as shown in the fourth row on the left side of \Fref{fig:qualitative}, the query may include a coordinate shift of the same building, or as in the second and fourth rows on the right side, the satellite query may reflect a different capture condition (e.g., season, weather, or time) than the ground truth, which can lower the Top-$k$ ranking. 

Nevertheless, even without any additional training or pairwise supervision, our framework successfully identifies the correct matching region (highlighted with a red bounding box) among the top retrieved results, visually demonstrating strong cross-view retrieval performance.

\section{Conclusion}

We present a training-free framework for Street-to-Satellite retrieval, integrating LLM-guided location inference, pretrained vision encoders, and geocoding APIs to generate semantically aligned satellite queries. Without training or auxiliary data, our framework achieves competitive performance using a single street-view image dataset. 

Extensive experiments show that our approach not only outperforms the existing contrastive learning-based methods but also sets a new state-of-the-art on the University-1652 benchmark in the Street-to-Satellite retrieval.
We further show that PCA-whitening consistently enhances retrieval robustness across various pretrained encoders in zero-shot settings, effectively bridging modality gaps. While our proposed method depends on the external APIs and LLMs, it enables scalable and automated construction of Street-to-Satellite image pairs, offering a practical foundation for future learning-based retrieval models. 

\section*{Acknowledgement}
This work was supported by the Research Program of the Electronics and Telecommunications Research Institute (ETRI), funded by [25ZD1120, Development of ICT Convergence Technology for Daegu-GyeongBuk Regional Industry].

This research was supported by the Regional Innovation System \& Education(RISE) Glocal 30 program through the Daegu RISE Center, funded by the Ministry of Education(MOE) and the Daegu, Republic of Korea.(2025-RISE-03-001)




{\small
\bibliographystyle{ieee_fullname}
\bibliography{egbib}

\begin{thebibliography}{10}\itemsep=-1pt

\bibitem{achiam2023gpt}
Josh Achiam, Steven Adler, Sandhini Agarwal, Lama Ahmad, Ilge Akkaya, Florencia~Leoni Aleman, Diogo Almeida, Janko Altenschmidt, Sam Altman, Shyamal Anadkat, et~al.
\newblock Gpt-4 technical report.
\newblock {\em arXiv preprint arXiv:2303.08774}, 2023.

\bibitem{babenko2015aggregating}
Artem Babenko and Victor Lempitsky.
\newblock Aggregating deep convolutional features for image retrieval.
\newblock {\em arXiv preprint arXiv:1510.07493}, 2015.

\bibitem{caron2020unsupervised}
Mathilde Caron, Ishan Misra, Julien Mairal, Priya Goyal, Piotr Bojanowski, and Armand Joulin.
\newblock Unsupervised learning of visual features by contrasting cluster assignments.
\newblock {\em Advances in neural information processing systems}, 33:9912--9924, 2020.

\bibitem{caron2021emerging}
Mathilde Caron, Hugo Touvron, Ishan Misra, Herv{\'e} J{\'e}gou, Julien Mairal, Piotr Bojanowski, and Armand Joulin.
\newblock Emerging properties in self-supervised vision transformers.
\newblock In {\em Proceedings of the IEEE/CVF international conference on computer vision}, pages 9650--9660, 2021.

\bibitem{deuser2023sample4geo}
Fabian Deuser, Konrad Habel, and Norbert Oswald.
\newblock Sample4geo: Hard negative sampling for cross-view geo-localisation.
\newblock In {\em Proceedings of the IEEE/CVF International Conference on Computer Vision}, pages 16847--16856, 2023.

\bibitem{dosovitskiy2020image}
Alexey Dosovitskiy, Lucas Beyer, Alexander Kolesnikov, Dirk Weissenborn, Xiaohua Zhai, Thomas Unterthiner, Mostafa Dehghani, Matthias Minderer, Georg Heigold, Sylvain Gelly, et~al.
\newblock An image is worth 16x16 words: Transformers for image recognition at scale.
\newblock {\em arXiv preprint arXiv:2010.11929}, 2020.

\bibitem{guo2025deepseek}
Daya Guo, Dejian Yang, Haowei Zhang, Junxiao Song, Ruoyu Zhang, Runxin Xu, Qihao Zhu, Shirong Ma, Peiyi Wang, Xiao Bi, et~al.
\newblock Deepseek-r1: Incentivizing reasoning capability in llms via reinforcement learning.
\newblock {\em arXiv preprint arXiv:2501.12948}, 2025.

\bibitem{jiang2023diego}
Albert~Q Jiang, Alexandre Sablayrolles, Arthur Mensch, Chris Bamford, and Devendra~Singh Chaplot.
\newblock Diego de las casas.
\newblock {\em Florian Bressand, Gianna Lengyel, Guillaume Lample, Lucile Saulnier, L{\'e}lio Renard Lavaud, Marie-Anne Lachaux, Pierre Stock, Teven Le Scao, Thibaut Lavril, Thomas Wang, Timoth{\'e}e Lacroix, and William El Sayed}, pages 50--72, 2023.

\bibitem{koch2015siamese}
Gregory Koch, Richard Zemel, Ruslan Salakhutdinov, et~al.
\newblock Siamese neural networks for one-shot image recognition.
\newblock In {\em ICML deep learning workshop}, volume~2, pages 1--30. Lille, 2015.

\bibitem{kordopatis2025ilias}
Giorgos Kordopatis-Zilos, Vladan Stojni{\'c}, Anna Manko, Pavel Suma, Nikolaos-Antonios Ypsilantis, Nikos Efthymiadis, Zakaria Laskar, Jiri Matas, Ondrej Chum, and Giorgos Tolias.
\newblock Ilias: Instance-level image retrieval at scale.
\newblock In {\em Proceedings of the Computer Vision and Pattern Recognition Conference}, pages 14777--14787, 2025.

\bibitem{liu2024remoteclip}
Fan Liu, Delong Chen, Zhangqingyun Guan, Xiaocong Zhou, Jiale Zhu, Qiaolin Ye, Liyong Fu, and Jun Zhou.
\newblock Remoteclip: A vision language foundation model for remote sensing.
\newblock {\em IEEE Transactions on Geoscience and Remote Sensing}, 2024.

\bibitem{liu2019lending}
Liu Liu and Hongdong Li.
\newblock Lending orientation to neural networks for cross-view geo-localization.
\newblock In {\em Proceedings of the IEEE/CVF conference on computer vision and pattern recognition}, pages 5624--5633, 2019.

\bibitem{liu2022swin}
Ze Liu, Han Hu, Yutong Lin, Zhuliang Yao, Zhenda Xie, Yixuan Wei, Jia Ning, Yue Cao, Zheng Zhang, Li Dong, et~al.
\newblock Swin transformer v2: Scaling up capacity and resolution.
\newblock In {\em Proceedings of the IEEE/CVF conference on computer vision and pattern recognition}, pages 12009--12019, 2022.

\bibitem{liu2021swin}
Ze Liu, Yutong Lin, Yue Cao, Han Hu, Yixuan Wei, Zheng Zhang, Stephen Lin, and Baining Guo.
\newblock Swin transformer: Hierarchical vision transformer using shifted windows.
\newblock In {\em Proceedings of the IEEE/CVF international conference on computer vision}, pages 10012--10022, 2021.

\bibitem{oquab2023dinov2}
Maxime Oquab, Timoth{\'e}e Darcet, Th{\'e}o Moutakanni, Huy Vo, Marc Szafraniec, Vasil Khalidov, Pierre Fernandez, Daniel Haziza, Francisco Massa, Alaaeldin El-Nouby, et~al.
\newblock Dinov2: Learning robust visual features without supervision.
\newblock {\em arXiv preprint arXiv:2304.07193}, 2023.

\bibitem{radenovic2018fine}
Filip Radenovi{\'c}, Giorgos Tolias, and Ond{\v{r}}ej Chum.
\newblock Fine-tuning cnn image retrieval with no human annotation.
\newblock {\em IEEE transactions on pattern analysis and machine intelligence}, 41(7):1655--1668, 2018.

\bibitem{radford2021learning}
Alec Radford, Jong~Wook Kim, Chris Hallacy, Aditya Ramesh, Gabriel Goh, Sandhini Agarwal, Girish Sastry, Amanda Askell, Pamela Mishkin, Jack Clark, et~al.
\newblock Learning transferable visual models from natural language supervision.
\newblock In {\em International conference on machine learning}, pages 8748--8763. PmLR, 2021.

\bibitem{regmi2019bridging}
Krishna Regmi and Mubarak Shah.
\newblock Bridging the domain gap for ground-to-aerial image matching.
\newblock In {\em Proceedings of the IEEE/CVF International Conference on Computer Vision}, pages 470--479, 2019.

\bibitem{sablayrolles2018spreading}
Alexandre Sablayrolles, Matthijs Douze, Cordelia Schmid, and Herv{\'e} J{\'e}gou.
\newblock Spreading vectors for similarity search.
\newblock {\em arXiv preprint arXiv:1806.03198}, 2018.

\bibitem{shen2023mccg}
Tianrui Shen, Yingmei Wei, Lai Kang, Shanshan Wan, and Yee-Hong Yang.
\newblock Mccg: A convnext-based multiple-classifier method for cross-view geo-localization.
\newblock {\em IEEE Transactions on Circuits and Systems for Video Technology}, 34(3):1456--1468, 2023.

\bibitem{shi2019spatial}
Yujiao Shi, Liu Liu, Xin Yu, and Hongdong Li.
\newblock Spatial-aware feature aggregation for image based cross-view geo-localization.
\newblock {\em Advances in Neural Information Processing Systems}, 32, 2019.

\bibitem{tian2020cross}
Yuxin Tian, Xueqing Deng, Yi Zhu, and Shawn Newsam.
\newblock Cross-time and orientation-invariant overhead image geolocalization using deep local features.
\newblock In {\em Proceedings of the IEEE/CVF Winter Conference on Applications of Computer Vision}, pages 2512--2520, 2020.

\bibitem{toker2021coming}
Aysim Toker, Qunjie Zhou, Maxim Maximov, and Laura Leal-Taix{\'e}.
\newblock Coming down to earth: Satellite-to-street view synthesis for geo-localization.
\newblock In {\em Proceedings of the IEEE/CVF Conference on Computer Vision and Pattern Recognition}, pages 6488--6497, 2021.

\bibitem{touvron2023llama}
Hugo Touvron, Thibaut Lavril, Gautier Izacard, Xavier Martinet, Marie-Anne Lachaux, Timoth{\'e}e Lacroix, Baptiste Rozi{\`e}re, Naman Goyal, Eric Hambro, Faisal Azhar, et~al.
\newblock Llama: Open and efficient foundation language models.
\newblock {\em arXiv preprint arXiv:2302.13971}, 2023.

\bibitem{tschannen2025siglip}
Michael Tschannen, Alexey Gritsenko, Xiao Wang, Muhammad~Ferjad Naeem, Ibrahim Alabdulmohsin, Nikhil Parthasarathy, Talfan Evans, Lucas Beyer, Ye Xia, Basil Mustafa, et~al.
\newblock Siglip 2: Multilingual vision-language encoders with improved semantic understanding, localization, and dense features.
\newblock {\em arXiv preprint arXiv:2502.14786}, 2025.

\bibitem{van2012heavenly}
Glen Van~Brummelen.
\newblock Heavenly mathematics: The forgotten art of spherical trigonometry.
\newblock 2012.

\bibitem{wang2021each}
Tingyu Wang, Zhedong Zheng, Chenggang Yan, Jiyong Zhang, Yaoqi Sun, Bolun Zheng, and Yi Yang.
\newblock Each part matters: Local patterns facilitate cross-view geo-localization.
\newblock {\em IEEE Transactions on Circuits and Systems for Video Technology}, 32(2):867--879, 2021.

\bibitem{wang2024learning}
Tingyu Wang, Zhedong Zheng, Zunjie Zhu, Yaoqi Sun, Chenggang Yan, and Yi Yang.
\newblock Learning cross-view geo-localization embeddings via dynamic weighted decorrelation regularization.
\newblock {\em IEEE Transactions on Geoscience and Remote Sensing}, 2024.

\bibitem{wang2024mfrgn}
Yuntao Wang, Jinpu Zhang, Ruonan Wei, Wenbo Gao, and Yuehuan Wang.
\newblock Mfrgn: Multi-scale feature representation generalization network for ground-to-aerial geo-localization.
\newblock In {\em Proceedings of the 32nd ACM International Conference on Multimedia}, pages 2574--2583, 2024.

\bibitem{workman2015wide}
Scott Workman, Richard Souvenir, and Nathan Jacobs.
\newblock Wide-area image geolocalization with aerial reference imagery.
\newblock In {\em Proceedings of the IEEE International Conference on Computer Vision}, pages 3961--3969, 2015.

\bibitem{ye2024cross}
Junyan Ye, Zhutao Lv, Weijia Li, Jinhua Yu, Haote Yang, Huaping Zhong, and Conghui He.
\newblock Cross-view image geo-localization with panorama-bev co-retrieval network.
\newblock In {\em European Conference on Computer Vision}, pages 74--90. Springer, 2024.

\bibitem{zeng2022geo}
Zelong Zeng, Zheng Wang, Fan Yang, and Shin’ichi Satoh.
\newblock Geo-localization via ground-to-satellite cross-view image retrieval.
\newblock {\em IEEE Transactions on Multimedia}, 25:2176--2188, 2022.

\bibitem{zhai2023sigmoid}
Xiaohua Zhai, Basil Mustafa, Alexander Kolesnikov, and Lucas Beyer.
\newblock Sigmoid loss for language image pre-training.
\newblock In {\em Proceedings of the IEEE/CVF international conference on computer vision}, pages 11975--11986, 2023.

\bibitem{zhang2024image}
Bo-Jian Zhang, Guang-Hai Liu, and Zuoyong Li.
\newblock Image retrieval using unsupervised prompt learning and regional attention.
\newblock {\em Expert Systems with Applications}, 247:122913, 2024.

\bibitem{zheng2020university}
Zhedong Zheng, Yunchao Wei, and Yi Yang.
\newblock University-1652: A multi-view multi-source benchmark for drone-based geo-localization.
\newblock In {\em Proceedings of the 28th ACM international conference on Multimedia}, pages 1395--1403, 2020.

\end{thebibliography}
}

\end{document}